\pdfoutput=1

\documentclass[11pt]{article}

\usepackage[preprint]{acl}

\usepackage{times}
\usepackage{latexsym}

\usepackage[T1]{fontenc}

\usepackage[utf8]{inputenc}
\usepackage{multirow}
\usepackage{microtype}

\usepackage{inconsolata}

\usepackage{graphicx}
\usepackage{amsmath}
\usepackage{color}
\usepackage{tikz}
\usetikzlibrary{shapes, decorations, arrows, calc, arrows.meta, fit,positioning}
\tikzset{
    -Latex, auto, node distance =1 cm and 1 cm,semithick,
    state/.style ={ellipse, draw, minimum width = 0.7 cm},
    point/.style = {circle, draw, inner sep=0.04cm,fill,node contents={}},
    bidirected/.style={Latex-Latex,dashed},
    el/.style = {inner sep=2pt, align=left, sloped}
}

\usepackage{amsthm}
\usepackage{amssymb}

\title{CR3G: Causal Reasoning for Patient-Centric Explanations in Radiology Report Generation}

\author{Satyam Kumar \\
  IIT Bombay \\
\texttt{satyam@minds.iitb.ac.in; learnsatyam@gmail.com} \\}

\usepackage{tabularx}

\usepackage[most]{tcolorbox}
\usepackage{lipsum}

\begin{document}
\maketitle
\begin{abstract}
Automatic chest X-ray report generation is an important area of research aimed at improving diagnostic accuracy and helping doctors make faster decisions. Current AI models are good at finding correlations (or patterns) in medical images. Still, they often struggle to understand the deeper cause-and-effect relationships between those patterns and a patient's condition. Causal inference is a powerful approach that goes beyond identifying patterns to uncover why certain findings in an X-ray relate to a specific diagnosis. In this paper, we will explore the prompt-driven framework Causal Reasoning for Patient-Centric Explanations in Radiology Report Generation (CR3G) that is applied to chest X-ray analysis to improve understanding of AI-generated reports by focusing on cause-and-effect relationships, reasoning and generate patient-centric explanation. The aim to enhance the quality of AI-driven diagnostics, making them more useful and trustworthy in clinical practice. CR3G has shown better causal relationship capability and explanation capability for 2 out of 5 abnormalities. 

\end{abstract}

\section{Introduction}
\texttt{Why do things happen the way they do?} This question is important in many fields,
from healthcare and finance to law. Humans can look for cause and
effect \citep{annurev:/content/journals/10.1146/annurev.psych.58.110405.085555}, leading to amazing inventions and progress. Often in healthcare, we observe the symptoms or outcomes of a medical condition, but the true causes remain hidden. For instance, a patient might have a persistent cough and shortness of breath. While these symptoms are visible to the doctor, the
underlying cause remains unseen, such as a bacterial infection, long-term smoking, or exposure to air pollutants. Doctors must deduce the hidden causes behind the visible
symptoms. In this case, diagnostic tests like chest X-rays can help uncover the cause, but
they do not always explain the full story. The same applies across healthcare, where we
often see the effects (symptoms). However, we must work to identify the underlying causes, such
as infections, lifestyle factors, or genetic predispositions, to provide effective treatment.
This is where causal inference comes into the picture. According to \citep{10.1093/biomet/82.4.669}, causal inference explores the hidden mechanisms behind our observations, providing a deeper
understanding of complex systems.  Causal inference helps us to distinguish between correlations (patterns that appear together) and true causation (events that directly cause other events). 

Causal reasoning\footnote{\url{https://oecs.mit.edu/pub/ee7y4opg/release/1}} involves determining how one event (the cause) leads to another event (the effect). To understand cause and effect better, comprehensive analysis must consider unseen causes beyond immediately visible. In medical imaging, such as chest X-rays, visible effects of lung disease must be contextualized with invisible causes like patient history and environmental exposures. This approach involves linking unseen reasons to observable data, enhancing our understanding of complex situations like medical conditions. By connecting hidden factors to visible evidence, we can develop a more complete and accurate interpretation of events or diagnoses \citep{10.5555/1642718}.  

Our contributions are:
\begin{itemize}

\item  Proposed \underline{C}ausal \underline{R}easoning for Patient-Centric Explanations in \underline{R}adiology \underline{R}eport
\underline{G}eneration (CR3G), a Large Language Model (LLM) for handling causal relationships, reasoning and providing a patient-centric explanation (Section \ref{subsec: Exp}).
\item Annotation of 500 frontal images of the existing Indiana University Chest X-ray (IUX-CXR) dataset \cite{10.1093/Jamia/ocv080} to include causal relationships, causal reasoning, and patient-centric explanations, all of which were verified by medical experts.
\item CR3G has shown better causal relationship capability and explanation capability for 2 out of 5 abnormalities (Section \ref{sec: Result}).
\end{itemize}

\section{Literature survey}
Radiology-GPT \citep{liu2023exploring}  is a specialized language model designed for radiology. It has been trained using a large dataset focused on radiology knowledge,
which helps it perform better than models like StableLM and LLaMA. GPT-4, a powerful language model, in its ability to generate radiology reports. By testing different prompting methods, the researchers \citep{liu2023exploring} found that GPT-4 either surpasses or matches these top models in various common radiology task
Radiology-Llama2 \citep{liu2023radiology}  is a specialized language model designed for
radiology, built on the Llama2 architecture. 

M4CXR architecture \citep{park2024m4cxr}, a multi-modal Large Language Model (LLM) designed to perform various tasks related to Chest X-ray report (CXR) interpretation, including medical report generation (MRG), visual grounding, and visual question answering (VQA). Additionally, like many deep-learning
models, M4CXR faces challenges with interpretability, making it difficult to discern how
the model arrives at its conclusions. 

Technical language in radiology reports also limits patient comprehension, an overlooked issue. The research by \cite{zhao2024x} presents Layman’s RRG framework, including a simplified dataset and semantic evaluation methods, enabling more robust assessments. Training on layman’s data improves semantic accuracy and reveals scaling advantages, addressing the technical complexity of radiology reports and enhancing patient understanding.
X-RGen \cite{chen2023s4m} proposes a holistic approach, emulating radiologists' workflows: observation, cross-region analysis, interpretation, and report writing. Integrating advanced feature extraction and multi-region clinical knowledge enhances pattern recognition and generates detailed reports across six anatomical regions, addressing existing limitations.
RaDialog \cite{pellegrini2023radialog} is a conversational AI model designed for radiology, combining image analysis with fine-tuned LLMs to generate and discuss accurate reports. 
\section{Experiments} 
\subsection{Dataset}
The experiment is conducted on the Indiana University Chest X-ray (IUX-CXR) dataset, which contains 7,470 chest X-ray images and 3,955 radiology reports \cite{10.1093/Jamia/ocv080}. The number of patients matches the number of reports; however, each patient is associated with two X-ray images: frontal and lateral. The data created is labeled with tags by humans and a computer program called Medical Text Indexer (MTI).

\subsection{Experimental Setup} \label{subsec: Exp}
\begin{figure*}[!htp]
    \centering
    \includegraphics[width=1.05\textwidth]{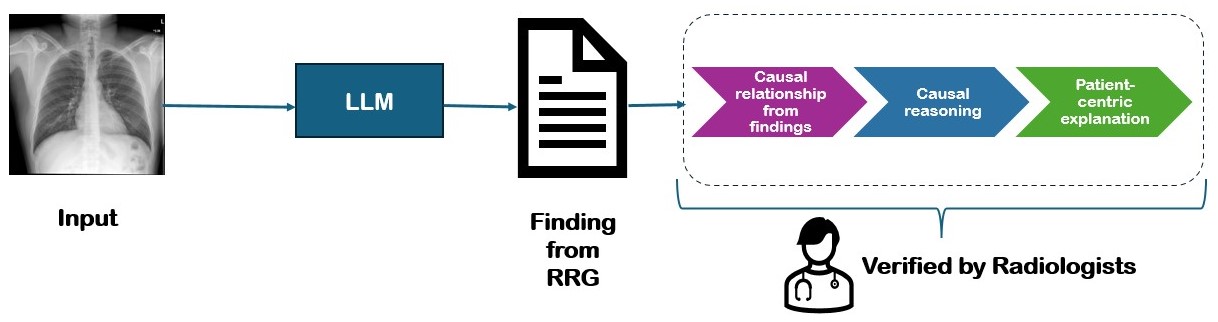}
    \caption{Pipeline for the Proposed methodology.}
    \label{fig:PM}
\end{figure*}

Figure \ref{fig:PM} shows the pipeline of the proposed methodology. The experiments were conducted on 2 NVIDIA A100 GPUs, each with 80GB VRAM. CR3G is built over the Large Language and Vision Assistant for BioMedicine (LLaVA-Med) \cite{li2024llava}. Firstly, LLaVA-Med is fine-tuned on MIMIC-CXR \cite{johnson2019mimic} to generate radiology reports. CLIP \cite{radford2021learning} encoder and GPT-4 \cite{achiam2023gpt} decoder is used in CR3G. The configurations included batch sizes of 64 and 96, learning rates of 0.01 and 0.05, and Adam optimizer with parameters $\beta_1 = 0.9$ and $\beta_2 = 0.099$. The dataset was divided into training, validation, and test sets with a split ratio of 7:1:2. Findings from the generated radiology reports were extracted using regular expressions. The prompts used for extracting causal relationships, causal reasoning, and patient-centric explanations are provided in Appendix \ref{appendix:a}. Based on the extracted findings, causal relationships, reasoning, and patient-centric explanations were identified. The patient-centric explanation simplifies complex medical findings into relatable terms.

   

\section{Result and Analysis} \label{sec: Result}
\begin{figure*}[!h]
    \centering
    \includegraphics[width=\textwidth, height=0.5\textwidth]{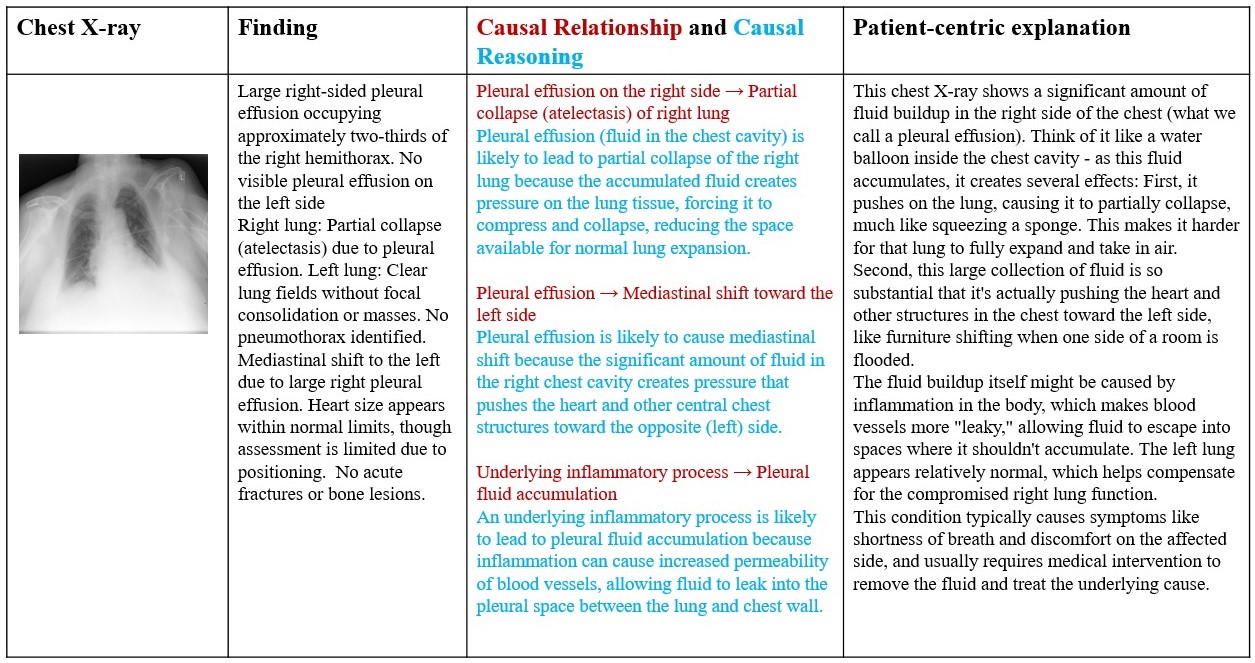}
    \caption{Causal relationships, reasoning, and patient-centric explanation based on findings obtained by CR3G.}
    \label{fig:PCE+CR-Result}
\end{figure*}

Figure \ref{fig:PCE+CR-Result} presents the chest X-ray image alongside findings, causal relationships (CR) with reasoning, and a patient-centric explanation (PCE) generated by CR3G. The analysis suggests that the given Chest X-ray image is of pleural effusion and its consequences, with only briefly mentioning heart size. Although the original tag indicates cardiomegaly, the causal relationship primarily addresses pleural effusion and atelectasis. According to the findings, the heart size appears within normal limits; however, positioning limits a conclusive assessment. If cardiomegaly were considered, potential causal relationships could involve chronic pressure changes or systemic conditions like heart failure contributing to pleural effusion. The reasoning might center on fluid accumulation due to impaired cardiac function, resulting in leakage into the pleural cavity. The explanation also highlights that inflammation might be causing the fluid buildup and reassures the patient that treatment can relieve symptoms like shortness of breath. to confirm cardiomegaly. 

Table \ref{tab:rating-table} shows the CR and PCE ratings obtained by Claude-3.5-sonnet and CR3G for 5 abnormalities found in Chest X-ray (CXR). It was found that CR3G generated better CR and PCE for Pneumothorax and Pleural effusion. The causal relationship (CR) rating is calculated on a scale of 1 to 5. A rating of 1 indicates no causal relationships are possible, and a rating of 5 means no changes are required. The PCE rating is rated on a scale of 1 to 5, where 1 indicates poor clarity, with explanations that are difficult to understand and contain many unexplained medical terms.  A rating of 5 represents exceptional clarity, with simple language and no use of jargon. Annotation guidelines of CR rating and justification with examples and PCE rating examples are provided in Appendix \ref{appendix:b}. For each abnormality, 100 CXR images are annotated with causal relationships, reasoning, and patient-centric explanations. The causal relationships, reasoning, and patient-centric explanations were verified by two medical experts in radiology. These ratings were averaged across 100 samples for each abnormality.

\begin{table*}[!h]
    \centering
    \begin{tabular}{|c|c|c|c|c|}
    \hline
    \multirow{2}{*}{\textbf{Abnormalities}} & \multicolumn{2}{|c|}{\textbf{Claude-3.5}} & \multicolumn{2}{|c|}{\textbf{CR3G}} \\ \cline{2-5}
         & \textbf{CR Correctness} & \textbf{PCE Rating} & \textbf{CR Correctness} & \textbf{PCE Rating} \\ \hline
         \textbf{Atelectasis} & 4 & 4.5 & 3 & 3 \\  \hline
         \textbf{Cardiomegaly} & 3.5 & 4.75 & 3.1 & 3.33\\ \hline
         \textbf{Pneumothorax} & 4.25 & 3.5 & 4.4 & 4.17\\ \hline
         \textbf{Pleural Effusion} & 3.75 & 3.38 & 4.33 & 3.67\\ \hline
         \textbf{Consolidation} & 5 & 4 & 3.2 & 3\\ \hline
    \end{tabular}
    \caption{Rating of Causal relationship (CR) correctness and Patient-centric explanation (PCE) by Claude-3.5 and CR3G.}
    \label{tab:rating-table}
\end{table*}

When analyzing causal relationships in chest X-rays (CXR), it is crucial to focus on true cause-and-effect connections that demonstrate how one finding directly leads to another. Some relationships from CR3G are merely associative or descriptive rather than causal. The presence of PICC lines indicating ongoing medical treatment (PICC lines present $\rightarrow$ Suggesting ongoing medical treatment), or the observation of Kerley B lines as a manifestation of interstitial pulmonary edema (Interstitial pulmonary edema $\rightarrow$ Bilateral Kerley B lines), are examples of such non-causal relationships (Refer to Figure \ref{fig:appendix-c-ex1}). Understanding this distinction is vital for accurate radiological interpretation, as it helps differentiate between direct physiological causes and effects versus radiographic signs or clinical associations. The observation that \textit{bilateral patchy airspace opacities suggest infection or inflammation} is actually a diagnostic interpretation rather than causation; the opacities are a manifestation or sign of the underlying condition, not a cause leading to an effect. 
\textit{Mild cardiomegaly indicating the potential for reduced cardiac output or heart failure} is a clinical association or risk assessment rather than a direct causal relationship. While cardiomegaly and heart failure are related, the relationship is more complex and bidirectional rather than purely causal. The Central venous catheter (CVC) in place leading to the potential risk of infection is also not causal but a risk assessment; a CVC does not directly cause infections, though it may increase the risk (Refer to Figure \ref{fig:appendix-c-ex2}). 
Similarly, \textit{ground-glass opacities leading to inflammatory process} reverses the actual pathophysiology; inflammation causes the ground-glass appearance, not vice versa. \textit{Inflammatory process leading to increased interstitial markings} is closer to a causal relationship. However, it still primarily describes how a disease process manifests radiographically rather than showing true causation.

For the causal relationship of tag \textit{Cardiomegaly}, Pleural effusion $\rightarrow$ Mediastinal shift toward the left side; Underlying inflammatory process $\rightarrow$ Pleural fluid accumulation, it is a case of indirect cause, and it is difficult for CR3G to create such causal relationships. Still, it was easier for the radiologist to create such a causal relationship. While an underlying inflammatory process, such as infection or malignancy, can lead to pleural fluid accumulation by increasing vascular permeability or disrupting pleural fluid absorption, the inflammation does not directly cause lung collapse or mediastinal shift. Instead, the pleural effusion that results from the inflammatory process exerts these mechanical effects. Therefore, the inflammatory process is indirectly related to atelectasis and mediastinal shift, leading to pleural effusion, which causes compression and displacement.

\section{Conclusion and Future Work}
We proposed CR3G framework that provides causal relationship, reasoning and patient-centric explanation.  We have also created an Annotation of 500 frontal images of the existing IUX-CXR dataset to include causal relationships, causal reasoning, and patient-centric explanations. CR3G has shown better causal relationship capability and explanation capability for 2 out of 5 abnormalities. True causal relationships in radiology should demonstrate clear, direct physiological mechanisms where one finding definitively leads to another. Causal reasoning is vital for interpreting chest X-rays (CXR) as it helps 
radiologists to identify the relationships between findings and underlying conditions,
enhancing diagnostic accuracy and treatment effectiveness. This understanding enables radiologists and clinicians to better plan appropriate interventions based on the underlying pathophysiology rather than just the observed findings. For future work, this approach should be extended to other chest X-ray datasets from different demographics, such as NIH-CXR14 \cite{wang2017chestx} and PadChest \cite{bustos2020padchest}. Additionally, we will explore counterfactual explanations with more robust causal relationships and causal reasoning frameworks.

\section{Limitations}
Causal inference typically requires large, high-quality datasets with diverse populations and detailed annotations, which are not easily available.
Causal reasoning through LLMs might still be opaque to clinicians if their reasoning process is not aligned with medical intuition or if they rely on abstract statistical metrics. Patient health can change rapidly; static causal models might not adapt well to such dynamics. Causes of findings may vary by geography (e.g., prevalence of tuberculosis) or population characteristics, which might not be adequately modeled. 
\section{Ethical consideration}
The authors of the IU Chest X-ray dataset employed appropriate techniques to de-identify the text reports. The data is anonymized, ensuring our model will not disclose any patient identity information.
\bibliography{custom}

\section*{Appendix}


    




\appendix
\section{Prompt Template}   \label{appendix:a}

 Generate a radiology report for the provided Chest X-ray image \{\{cxr\_image\}\}.\\
Task 1: Identify and describe causal relationships based on the findings observed in the image. Use the following format for relationships:   \{cause\} -> \{effect\}

Example format:
\begin{itemize}
    \item Markedly elevated right hemidiaphragm -> Atelectasis
    \item Atelectasis, Consolidation, Pleural effusion  -> Minimal blunting of posterior costophrenic angle.
\end{itemize}
    
Create causal reasoning to explain each of the relationships from Task 1. Explanations should be in the format: \textit{Cause\_Finding} is likely to lead to \textit{Effect\_Finding} because \textit{Reasoning}.

Example format:
A markedly elevated right hemidiaphragm is likely to lead to atelectasis due to restricted lung expansion on the right side, limiting airflow and causing the collapse of lung tissue.

Provide a simplified explanation that combines the findings, causal relationships, and reasoning in a way that is understandable to non-experts (e.g., patients or newly joined doctors).


\section {Generation of non-causal relationship examples}   \label{appendix:c}

\begin{figure}[!h]
    \includegraphics[width=0.5\textwidth, height=0.8\linewidth]{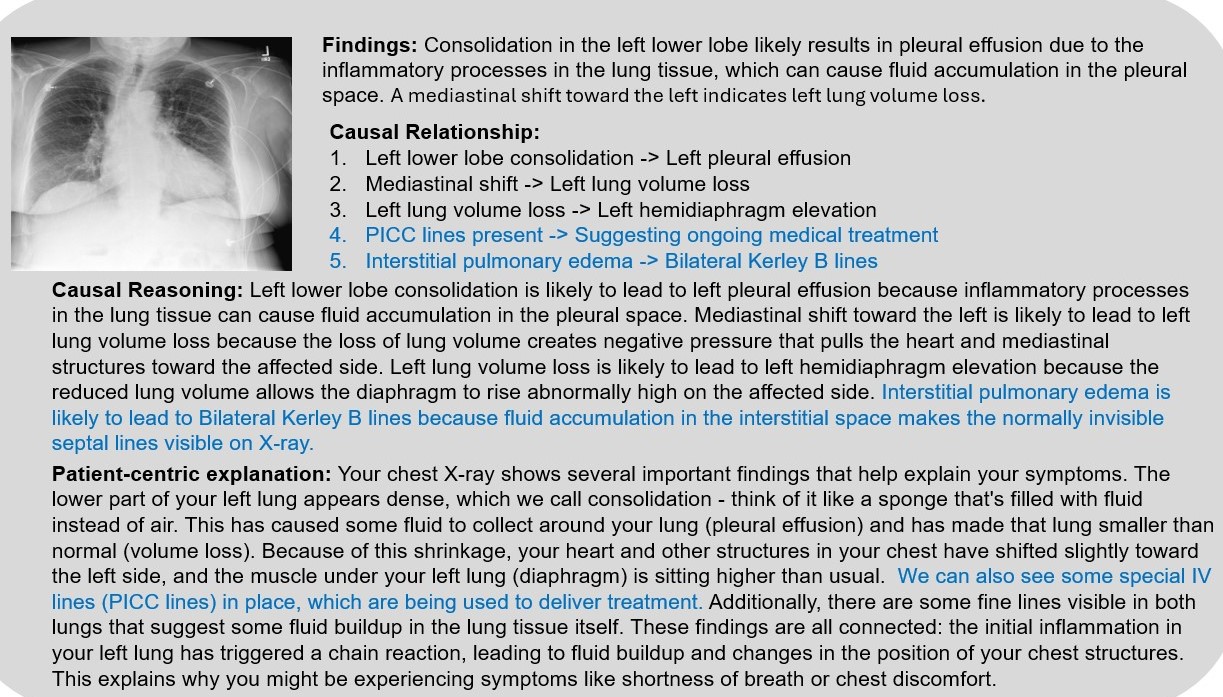}
    \caption{Association-based relationships, not causal relationships, are depicted in blue text}
    \label{fig:appendix-c-ex1}
\end{figure}

\begin{figure}[!h]
    \includegraphics[width=0.5\textwidth, height=0.8\linewidth]{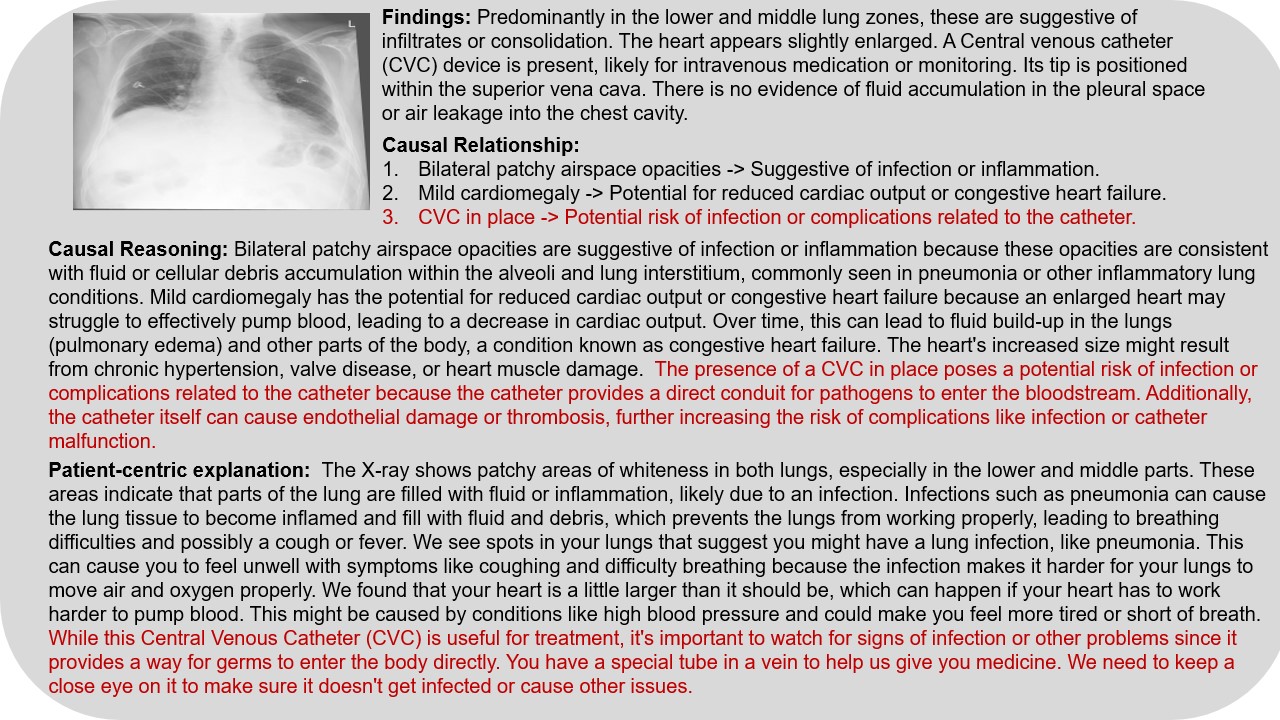}
    \caption{Potential risk of infection is not causal but a risk assessment, indicated by red text.}
    \label{fig:appendix-c-ex2}
\end{figure}

\begin{figure}[!h]
    \includegraphics[width=0.5\textwidth, height=0.8\linewidth]{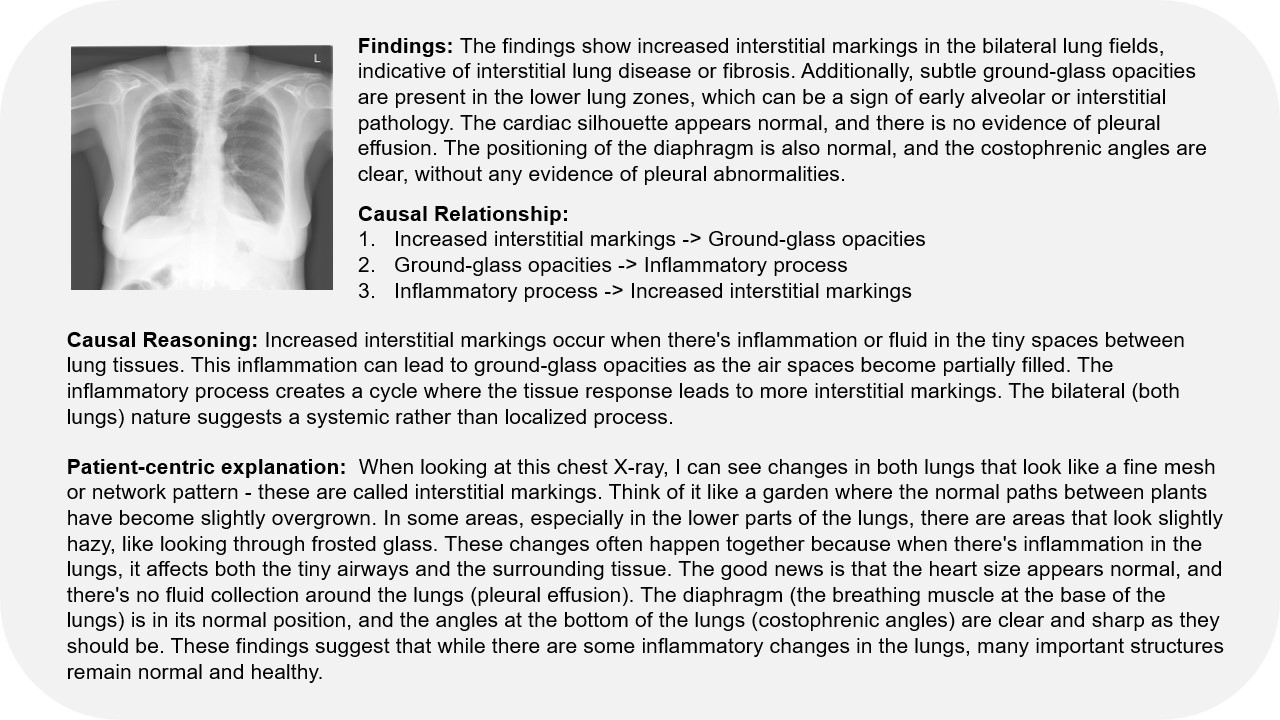}
    \caption{Presence of cycle in the causal relationship.}
    \label{fig:appendix-c-ex3}
\end{figure}

In this section, a few samples where CR3G generated non-causal relationships. Causal relationships 4 and 5 from Figure \ref{fig:appendix-c-ex1} show that these are association-based relationships that must be removed from the causal relationship, reasoning, and PCE. Similarly, in Figure \ref{fig:appendix-c-ex2}, causal relationship, reasoning, and PCE must be removed as those are potential risks of infection that are not causal. Figure \ref{fig:appendix-c-ex3} shows presence of cycle in causal relationship, one of the causal relationship is not required. Accordingly the changes have to made in causal reasoning and PCE.


\section {Annotation guidelines} \label{appendix:b}
\appendix
1. Causal Relationship (CR) Rating Guidelines

The causal relationship (CR) rating evaluates the accuracy and completeness of causal relationships described in the radiology report. The rating is calculated on a scale of 1 to 5:
\begin{itemize}
\item Rating 1: No causal relationships are identified or possible within the findings.

\item Rating 2: Few causal relationships are described, with significant inaccuracies in reasoning.

\item Rating 3: Some causal relationships are accurately identified, but the reasoning may be incomplete or partially incorrect.

\item Rating 4: Most causal relationships are accurately identified and well-reasoned, with only minor inaccuracies or omissions.

\item Rating 5: All causal relationships are accurately identified.
\end{itemize}

\textbf{Examples of CR Ratings and Justifications}
\begin{itemize}
\item    Atelectasis  $\rightarrow$ Low lung volumes (Rating 1)
\begin{itemize}
\item \textbf{Justification:}  Direction of the causal relationship is wrong; low lung volumes $\rightarrow$ Atelectasis and also it is more of association relationship
\end{itemize}
\item Right Hemidiaphragm Elevation $\rightarrow$ Consolidation (Rating : 2)\\
 \begin{itemize}
     \item \textbf{Justification:}  Consolidation, typically related to infection or inflammation, is unlikely to be directly caused by right hemidiaphragm elevation. The diaphragm's elevation might be an effect of consolidation (e.g., due to inflammation or pleural effusion), but not a cause of it. This causal relationship is association but not causal relationship.
 \end{itemize}
   \item Bilateral patchy airspace opacities $\rightarrow$ Suggestive of infection or inflammation. (Rating: 3)\\
 \begin{itemize}
     \item \textbf{Justification:} Bilateral patchy airspace opacities are commonly seen in infections (such as pneumonia) or inflammation (like interstitial lung disease), so the causal relationship is reasonable. However, other conditions might also cause similar opacities, so the relationship is plausible but not definitive.
 \end{itemize}
 \item Right Hemidiaphragm Elevation $\rightarrow$ Mediastinal Shift (Rating: 4)
 \begin{itemize}
     \item \textbf{Justification:} When the diaphragm elevates on the right side, it can cause a shift in the mediastinal structures, especially if there's volume loss or other factors leading to pressure imbalances. The relationship is clear and practically possible but could be better explained with specific details such as the degree of diaphragm elevation
 \end{itemize}
 
 \end{itemize}

2. Patient-centric Explanation (PCE) Rating Guidelines
The PCE rating assesses the clarity and accessibility of the explanation provided for non-experts. The rating is based on a scale of 1 to 5:
\begin{itemize}
    \item Rating 1: Poor clarity, with explanations that are difficult to understand and contain many unexplained medical terms.

\item Rating 2: Somewhat clear explanations, but still containing medical jargon that may confuse non-experts.

\item Rating 3: Mostly clear explanations, although some medical terms require further clarification or simplification.

\item Rating 4: Clear and easy-to-understand explanations, with only minor use of medical jargon.

\item Rating 5: Exceptional clarity, using simple language and avoiding jargon entirely.
\end{itemize}

\textbf{Examples of PCE Rating}
\begin{itemize}
\item \textbf{(Rating: 1)} Left lower lobe consolidation is likely to lead to left pleural effusion because inflammatory processes in the lung tissue can cause fluid accumulation in the pleural space. 
\item  \textbf{(Rating: 2)} When looking at this chest X-ray, the changes in both lungs that look like a fine mesh or network pattern - these are called interstitial markings.  Some areas, especially in the lower parts of the lungs, look slightly hazy, like looking through frosted glass. 
\item \textbf{(Rating: 3)} The chest X-ray shows that the right side of your diaphragm, which is a muscle that helps you breathe, is higher than normal. This has caused the middle part of your chest, called the mediastinum, to shift to the right. Because of this change, part of your right lung is not fully expanding, known as atelectasis.  Additionally, a small amount of fluid around your right lung can be seen as a blunting of the angle where your lung meets your ribs. This fluid can press on your lungs, making it harder for you to breathe deeply.
\item \textbf{(Rating: 4)} The lower part of your left lung appears dense, which we call consolidation - think of it like a sponge filled with fluid instead of air.  These findings are all connected: the initial inflammation in your left lung has triggered a chain reaction, leading to fluid buildup and changes in the position of your chest structures. This explains why the patient might be experiencing symptoms like shortness of breath or chest discomfort.
\item \textbf{(Rating: 5)} The chest X-ray shows a few things that need attention. There is some cloudiness or haziness at the bottom of both lungs, more so on the right. This could mean there is an inflammation or fluid buildup in those areas or that some parts of your lung are not inflating as well as they should. It can also be seen that the muscle separating your chest from your abdomen (the diaphragm) is a little higher than normal on the right side. This could be because of the issues in your lung, or it could mean there is some fluid around your lung. The spot where your ribs meet your diaphragm on the right looks a bit rounded, which could be another sign of fluid. This is called a small pleural effusion. A thin tube is placed in a large vein leading to your heart, which is likely there to help deliver medications or fluids. Your heart also appears slightly larger than expected. In simpler terms, you might have some fluid or inflammation in the lower parts of your lungs, particularly on the right side. This could be causing some of the other findings we're seeing on the X-ray. We'll likely need to do some further tests to understand what is causing these changes and to decide on the best course of treatment.
 \end{itemize}



\end{document}